\documentclass[a4paper]{article}

\usepackage{INTERSPEECH2022}
\usepackage{url}
\usepackage{multirow,multicol}
\usepackage{bbding}
\usepackage{booktabs}
\usepackage{graphicx}
\usepackage{makecell}

\title{Blockwise Streaming Transformer for Spoken Language Understanding and Simultaneous Speech Translation}
\name{Keqi Deng$^1$, Shinji Watanabe$^2$, Jiatong Shi$^2$, Siddhant Arora$^2$ \thanks{$^*$ We will release this streaming ST/SLU publicly available through the ESPnet open source toolkit.}}
\address{
  $^1$University of Chinese Academy of Sciences, China\\
  $^2$Carnegie Mellon University, USA}
\email{dengkeqi20@mails.ucas.ac.cn, shinjiw@cmu.edu, \{jiatongs, siddhana\}@cs.cmu.edu}

\begin{document}

\maketitle
\begin{abstract}
Although Transformers have gained success in several speech processing tasks like spoken language understanding (SLU) and speech translation (ST), achieving online processing while keeping competitive performance is still essential for real-world interaction. In this paper, we take the first step on streaming SLU and simultaneous ST using a blockwise streaming Transformer, which is based on contextual block processing and blockwise synchronous beam search. 
Furthermore, we design an automatic speech recognition (ASR)-based intermediate loss regularization for the streaming SLU task to improve the classification performance further. As for the simultaneous ST task, we propose a cross-lingual encoding method, 
which employs a CTC branch optimized with target language translations. In addition, the CTC translation output is also used to refine the search space with CTC prefix score, achieving joint CTC/attention simultaneous translation for the first time. 
Experiments for SLU are conducted on FSC and SLURP corpora, while the ST task is evaluated on Fisher-CallHome Spanish and MuST-C En-De corpora. 
Experimental results show that the blockwise streaming Transformer achieves competitive results compared to offline models, especially with our proposed methods that further yield a 2.4$\%$ accuracy gain on the SLU task and a 4.3 BLEU gain on the ST task over streaming baselines.

\end{abstract}

\section{Introduction}
In the last decade, deep learning has greatly promoted the
development of several speech processing tasks like spoken language understanding (SLU) \cite{8639043} and speech translation (ST) \cite{9054585}.
SLU task aims to extract structured semantic representations from speech signals \cite{kuo20_interspeech,Potdar2021ASE}. Conventional cascaded SLU systems consist of an automatic speech recognition (ASR) module as well as a downstream natural language understanding (NLU) module \cite{article}.
On the other hand, end-to-end (E2E) SLU systems directly extract users’ intentions from input speech to avoid error propagation seen in the above cascaded method \cite{8639043, arora2021espnet}. Similarly, E2E ST systems directly translate source language speech into target language text and have advantages such as lower latency, smaller model size, and less error compounding over cascaded ST \cite{DBLP:conf/aaai/LiuZXZHWWZ20,9687894}. 
However, for real-world human-computer interactions, it is still essential to make the systems online while keeping competitive performance.

Online systems effectively reduce the processing latency by in-time responses before consuming the full input speech \cite{Potdar2021ASE}. 
For streaming SLU task, \cite{Potdar2021ASE} takes the first step to achieve an E2E streaming SLU model based on connectionist temporal classification (CTC) objective \cite{10.5555/3044805.3045089}. \cite{cao21c_interspeech} further employs an adapted version, named connectionist temporal localization, for online SLU. These approaches identify intent when sufficient evidence has been accumulated, without waiting until the end of the utterance.
As for the simultaneous ST (SST) task, 
recent works can be divided into two categories: fixed policy and flexible policy \cite{ma-etal-2020-simulmt}. 
Several works employ fixed policy to SST \cite{Ma2021StreamingSS} by adapting the text-based wait-k strategy \cite{ma-etal-2019-stacl}, while we also observe works with flexible policies through monotonic attention \cite{DBLP:journals/corr/abs-2110-15729} and other variants \cite{xutai2021mono, DBLP:conf/iclr/ChiuR18}.

However, unlike simultaneous text translation, whose input is already segmented
into words, the wait-k strategy faces a challenge of computing the number of valid tokens for a source speech when applied to the SST task \cite{chen-etal-2021-direct}. And flexible policy methods like MoChA \cite{DBLP:conf/iclr/ChiuR18} significantly degrade translation quality and make it difficult to keep an acceptable latency \cite{9383517}. In addition, the current decoding process of ST is mostly based on attention-based inference, which suffers from poor text length prediction \cite{9398531}.
Meanwhile, on streaming SLU tasks, although the CTC model has been proved effective, online systems based on popular encoder-decoder structures are still worth exploring.

In this paper, 
we propose to use blockwise streaming Transformer \cite{9383517}, which has been proved to outperform prior methods like MoChA in the ASR task, for streaming SLU and simultaneous ST systems.
Inspired by \cite{Encoder},
we further design an ASR-based intermediate loss to help the model better converge for streaming SLU. As for simultaneous ST, we propose a cross-lingual encoding (CLE) method by injecting a CTC objective between encoder outputs and target translations. In addition, the CTC-based alignment is also used to refine the search space via considering the CTC prefix scores, by which we achieve a joint CTC/attention simultaneous translation for the first time. 
Experimental results show that the blockwise streaming Transformer originally developed for ASR can achieve competitive results on SLU and ST tasks. 
And our proposed methods can further yield a 2.4$\%$ accuracy gain on the streaming SLU task and a 4.3 BLEU gain on the simultaneous ST task.

\section{Blockwise Streaming Transformer}
\label{streaming}
\begin{figure}[t!]
    \centering
    \vspace{-0.3cm}
    \setlength{\belowcaptionskip}{-0.5cm}
    \includegraphics[width=0.90\linewidth]{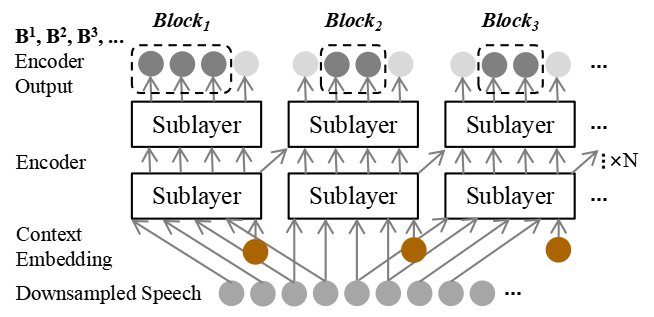}
    \caption{Illustration of the contextual block processing method.}
    \label{fig:streaming}
\end{figure}

Block processing is an effective way to make the Transformer encoder online \cite{9054476,9003749,9414560}. As shown in Fig.~\ref{fig:streaming},
\cite{9003749} introduces a context inheritance mechanism to utilize richer contextual information. Previous context is encoded into context embedding, which is calculated for each block at each sublayer
and then
handed over to the next sublayer.
We denote the $i$-th block of encoded feature as ${\rm\mathbf{B}}^i=({\boldsymbol{B}}^i_1, \cdot\cdot\cdot,{\boldsymbol{B}}^i_{T})$, where $T$ is block size.

To achieve streaming decoding, \cite{9383517} further proposes a blockwise synchronous beam search based on the contextual block processing \cite{9003749}.
The decoder predicts next word $y_j$ based on previous output $y_{1:j-1}$ and $b$ block feature ${\rm\mathbf{B}}^{1:b}$. 
In addition, a CTC prefix score is also computed based on ${\rm\mathbf{B}}^{1:b}$ to achieve joint CTC/attention decoding \cite{9383517, CTC-Attention-ACL-2017}.
When a hypothesis contains an end token or a repetition, this prediction is regarded as unreliable with insufficient $b$ blocks, and then the decoder waits for the next block to be encoded \cite{9383517}.

\section{Proposed Method}
In this paper, we propose to use a blockwise streaming Transformer for the streaming SLU and simultaneous ST tasks. 
\subsection{Streaming spoken language understanding}
E2E SLU combines ASR and NLU into one task \cite{Potdar2021ASE}, thus requiring both acoustic and semantic understanding \cite{arora2021espnet}. 
Prior work observes that adding auxiliary ASR objectives by training models to predict both intent and transcript can improve the performance of SLU systems \cite{arora2021espnet}.\footnote{This method adds intent right before the transcript, and then uses the system as an ASR model.}

\begin{figure}[h]
    \centering
    \vspace{-0.3cm}
    \includegraphics[width=0.72\linewidth]{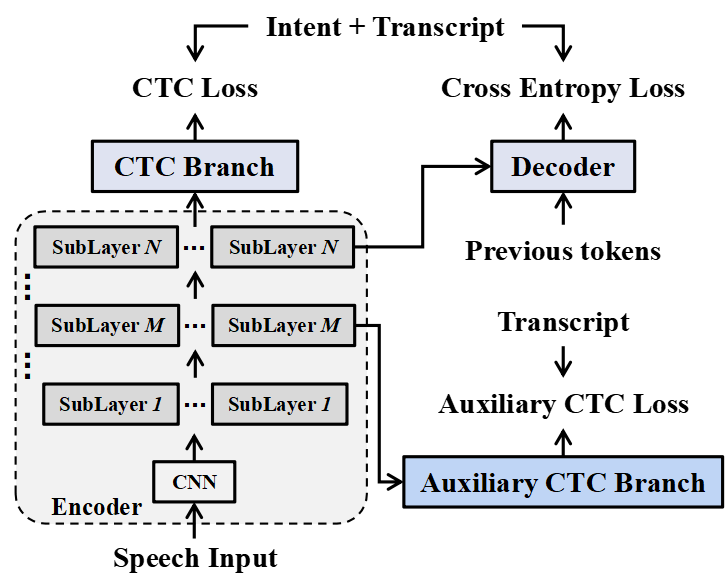}
    \setlength{\belowcaptionskip}{-0.5cm}
    \caption{Illustration of the ASR-based intermediate loss regularization, which is denoted as auxiliary CTC loss.}
    \label{fig:slu}
\end{figure}

However, unlike the ASR token, the intent has no corresponding speech frames, which makes it hard for the monotonic alignment model like CTC to learn the target that contains both intent and ASR transcripts. Therefore, we argue that a model should first distinguish words before learning to understand intents.
Under the framework of blockwise streaming Transformer,
we use an ASR-based intermediate loss regularization method to promote the learning process, which is shown in Fig.~\ref{fig:slu}. We apply an extra CTC branch to the $M$-th encoder layer and an auxiliary CTC loss is computed with ASR transcripts as the target. In this way, lower
layers of the encoder are encouraged to distinguish different tokens, while the higher encoder layers try to understand semantics for intent classification.
The final training objective $\mathcal{L}_{\rm{mtl}}^{\text{slu}}$ is calculated as follows:
\begin{equation}
    \mathcal{L}_{\text{mtl}}^{\text{slu}}=\lambda (\mathcal{L}_{\rm{ctc}}+\mathcal{L}_{\rm{ctc}}^{\text{aux}}) + (1\!-\!\lambda)\mathcal{L}_{\rm{ce}}, \label{mot}
\end{equation}
\begin{table*}[t]
  \caption{BLEU and average lagging (AL) (ms) of different ST systems on Fisher-CallHome Spanish corpus. Details about the block size and cross-lingual encoding (CLE) are respectively introduced in Section~\ref{streaming} and Section~\ref{subsec:CLE}. }
  \label{tab:fish}
  \centering
  \setlength{\tabcolsep}{1.5mm}
  \begin{tabular}{l c c c | c c c | c c|c}
    \Xhline{3\arrayrulewidth}
    \multirow{2}{*}{Model}&
    {Cross-lingual} &\multirow{2}{*}{Block Size}&
    \multirow{2}{*}{Decoding Style}&\multicolumn{3}{c|}{Fisher}&\multicolumn{2}{c|}{CallHome}&\multirow{2}{*}{AL} \\
    \cline{5-9}
     & {Encoding}& & &dev &dev2 &test & devtest&evltest&  \\
    \hline
    Offline Transformer&\XSolidBrush&\XSolidBrush& Attention-based beam seach&
    44.5&45.3&44.6&15&14.4&\multirow{2}{*}{---}\\
    { }{ } + ASR encoder init.&\XSolidBrush&\XSolidBrush&Attention-based beam seach&48.2&48.1&48.0&16.6&16.3&\\
    \hline
    Streaming Transformer&\XSolidBrush &20& Attention-based beam search&40.6
    &41.5&40.7&11.9&10.9&3298\\
     { }{ } + ASR encoder init.&\XSolidBrush &20& Attention-based beam search&
     44.4&45.4&44.3&14.2&13.6&3236\\
     Streaming Transformer&\Checkmark&20&Attention-based beam search&
     40.9&40.9&40.7&12.3&11.8&3261\\
    { }{ } + ASR encoder init.&\Checkmark&20&Attention-based beam search&
    45.2&45.4&45.4&14.2&13.8&\textbf{3232}\\
    Streaming Transformer&\Checkmark&20&CTC/attention joint translation&
    43.6&44.1&43.4&13.5&13.5&3319\\
    { }{ } + ASR encoder init.&\Checkmark&20&CTC/attention joint translation&
    \textbf{47.4}&\textbf{47.9}&\textbf{47.7}&\textbf{15.0}&\textbf{15.2}&3257\\
    \hline
    Streaming Transformer&\XSolidBrush &40& Attention-based beam search&
    41.0&42.0&41.1&12.8&12.4&\textbf{3361}\\
     { }{ } + ASR encoder init.&\XSolidBrush &40& Attention-based beam search&
     44.6&45.7&45.2&14.2&13.9&3404\\
     Streaming Transformer&\Checkmark&40&Attention-based beam search&
     41.1&41.2&41.2&12.3&12.2&3376\\
    { }{ } + ASR encoder init.&\Checkmark&40&Attention-based beam search&
    45.4&46.5&45.3&14.8&14.4&3416\\
    Streaming Transformer&\Checkmark&40&CTC/attention joint translation&
    43.6&43.8&43.6&13.3&13.4&3426\\
    { }{ } + ASR encoder init.&\Checkmark&40&CTC/attention joint translation&
    \textbf{47.9}&\textbf{48.2}&\textbf{47.7}&\textbf{15.5}&\textbf{15.3}&3434\\
    \Xhline{3\arrayrulewidth}
  \end{tabular}
\end{table*}
where $\lambda \in [0,1]$, $\mathcal{L}_{\rm{ctc}}$ is the main CTC loss,
$\mathcal{L}_{\rm{ctc}}^{\text{aux}}$ denotes the auxiliary loss, and $\mathcal{L}_{\rm{ce}}$ represents the cross-entropy (CE) loss. This extra CTC branch is discarded during inference thus does not break the online algorithms of streaming Transformer.

\subsection{Simultaneous speech translation}
E2E ST combines ASR and machine translation (MT) into a single task \cite{Le2020DualdecoderTF}. The optimization of E2E ST models can be more difficult than individually training ASR and MT models \cite{inaguma-etal-2020-espnet}. Multi-task learning and pre-training methods from ASR tasks are always used to alleviate the problem \cite{inaguma-etal-2020-espnet,9003832}.

Current works on ST mostly rely on attention-based decoding, which 
results in poor generation due to wrong text length \cite{9398531}. In this paper, we propose a cross-lingual encoding method and thus achieve joint CTC/attention simultaneous translation for the first time. 
It should be noted that our proposed methods do not break our online algorithms thanks to flexible blockwise streaming Transformer algorithms.

\subsubsection{Cross-lingual encoding (CLE)}
\label{subsec:CLE}
Our proposed cross-lingual encoding (CLE) method employs 
a CTC branch optimized with target language translations.
Although the CTC's success on the ST task is counter-intuitive due to its monotonic property, previous works \cite{DBLP:conf/acl/2021f, 9688157} have proved that Transformer with CTC as the objective has reordering capability \cite{DBLP:conf/acl/2021f}. As for simultaneous ST task,
the reordering capability of the Transformer is somewhat limited, but 
the context inheritance mechanism of blockwise streaming Transformer is capable of encoding the context of previously processed blocks \cite{9003749}. Therefore, we believe that CTC is still feasible and helps to refine the search space during decoding \footnote{Strictly speaking, CLE may have limited effect for language pairs with very serious reordering issue, but in this case, achieving simultaneous ST itself is a very hard topic. Therefore, we believe that our proposed CLE is feasible for selected simultaneous ST scenarios like English to German.}.

During training, with the target language translations as the learning target,
we calculate a CTC loss from the CTC branch applied after the encoder and a CE loss from the decoder. In addition, to help the encoder converge better, we also apply an ASR-based intermediate CTC loss for ASR multi-task learning. The final training objective $\mathcal{L}_{\text{mtl}}^{\text{st}}$ is computed as follows, where $\mathcal{L}_{\rm{ctc}}$ and $\mathcal{L}_{\rm{ce}}$ are for ST task while $\mathcal{L}_{\rm{ctc}}^{\text{aux}}$ is for the auxiliary ASR task. $\gamma$ and $\beta$ are weights of CTC loss for ST and ASR, respectively.
\begin{equation}
    \mathcal{L}_{\text{mtl}}^{\text{st}}=(1\!-\!\gamma)((1\!-\!\beta)\mathcal{L}_{\rm{ce}}+\beta\mathcal{L}_{\rm{ctc}})+\gamma\mathcal{L}_{\rm{ctc}}^{\text{aux}} \label{mot2}
\end{equation}
\subsubsection{Joint CTC/attention simultaneous translation}
Attention-based encoder-decoder (AED) has become the most popular structure for the E2E ST task \cite{tang-etal-2021-improving,DBLP:conf/acl/HuangWX21}. 
The AED system solves the ST task as a sequence mapping and
utilizes an attention mechanism \cite{Vaswani2017} to achieve alignments between acoustic inputs and translated tokens. However, the attention mechanism 
mainly relies on the dependency between decoder states to decide whether to stop, thus
having a flaw in poor generation performance due to wrong text length \cite{9398531}. Furthermore, in the E2E ST task, the length mismatch between acoustic input and translated tokens is distinct and varies greatly from case to case, making it harder to track the attention-based alignments \cite{8068205}.

\begin{figure}[h]
    \centering
    \vspace{-0.1cm}
    \setlength{\belowcaptionskip}{-0.3cm}
    \includegraphics[width=0.78\linewidth]{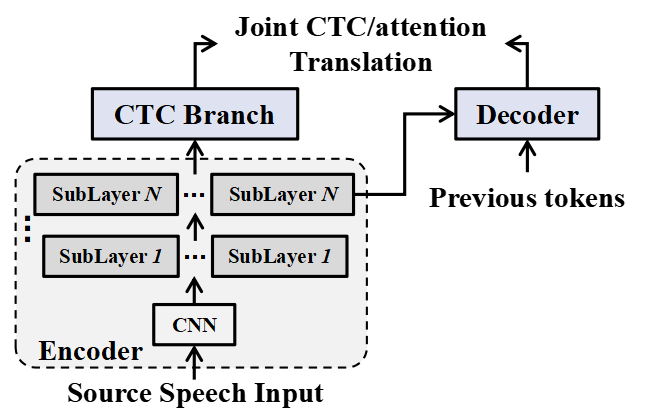}
    \caption{Illustration of the proposed joint CTC/attention simultaneous translation.}
    \label{fig:st}
\end{figure}

Therefore, we aim to utilize the CTC output to refine the search space and eliminate
irregular alignments during decoding \cite{CTC-Attention-ACL-2017}. Following the hybrid CTC/attention method \cite{8068205} and the blockwise synchronous beam search \cite{9383517} developed for the ASR task, we achieve a joint CTC/attention simultaneous translation based on our proposed CLE method and it is shown in Fig.~\ref{fig:st}.
The simultaneous ST process considers both the attention-based decoder's beam search scores $S_{\text{att}}$ and CTC's prefix scores $S_{\rm{ctc}}$ to predict the $j$-th token:
\begin{eqnarray}
    S_{\rm{ctc}}&=&\log p_{\rm ctc}(y_{j}|y_{1:j-1}, {\rm\mathbf{B}}^{1:b}), \\
    S&=&\mu S_{{\rm{ctc}}} + (1\!-\!\mu)S_{{\text{att}}}, \label{com3}
\end{eqnarray}
where $y_{1:j-1}$ is previous output, ${\rm\mathbf{B}}^{1:b}$ denotes $b$ block feature, and
$\mu$ represents the weights of CTC's scores.
\section{Experiments}
\subsection{Corpus}
We evaluate our SLU systems on Fluent Speech Commands (FSC) \cite{lugosch19_interspeech} and Spoken Language Understanding Resource Package (SLURP) \cite{slurp} corpora.
We use the official train, dev, and test sets for all our experiments.

As for the ST task, the experiments are conducted on Fisher-CallHome Spanish \cite{post-etal-2013-improved} and Must-C En-De \cite{di-gangi-etal-2019-must} corpora. We report case-insensitive BLEU.
Fisher-CallHome Spanish corpus 
consists of 170 hours Spanish speech, the Spanish transcripts, and the English translations. Must-C En-De contains English speech collected from TED talks, the English transcripts, and the German translations.

\subsection{Model descriptions}
We used ESPnet2 toolkit \cite{Watanabe2018ESPnet} to build the models.
For acoustic input, we employ 80-dimensional filter banks. For text output of the SLU task, we use 127 word units with 31 intents for FSC and 498 word units with 69 intents for SLURP. As for the ST task, for both source and target language, we employ subwords based on BPE \cite{gage1994} with 500 and 4000 sizes for Fisher-CallHome Spanish and Must-C En-De corpora, respectively.

We develop offline Transformer baseline models following the recipe of ESPnet2 \cite{Watanabe2018ESPnet}, which employs multi-task learning from ASR task (i.e. an ASR CTC branch applied after the encoder).
Our streaming Transformer models also have a 12-layer encode and a 6-layer decoder, in which the attention dimension, feed-forward dimension, and attention heads are kept the same as the offline models. 
And the ASR-based intermediate loss is computed from the 8-$th$ encoder layer.
For block processing \cite{9383517}, we keep both hop size and look-ahead size accounting for $20\%$ of the block size (after down-sampling).
As for Must-C En-De \footnote{The ESPnet2 example on this corpus is not yet open source in the current version of ESPnet.}, both the streaming and offline Transformer models have 256 attention dimensions, 2048 feed-forward dimensions, and 4 heads.
During streaming decoding, our implementation is based on 640 ms simulated chunk \footnote{\url{https://github.com/espnet/espnet/blob/master/espnet2/bin/asr_inference_streaming.py}}. For simultaneous ST task, we evaluate
the latency with average lagging (AL) defined in \cite{ma-etal-2020-simulmt} on the evltest of CallHome set or the tst-COMMON of Must-C.
As for streaming SLU task, we report endpoint latency (EP) \cite{chang17_interspeech} using a TITAN RTX GPU.

To prevent overfitting, we adopt a model averaging method. SpecAugment \cite{Park2019} is also used. $\lambda$ in Eq.~\ref{mot} is set to 0.3, while  $\beta$ and $\gamma$ in Eq.~\ref{mot2} are also set to 0.3.
During decoding, for the SLU task, the CTC weight is set to 0.5; as for the ST task, the CTC weight $\mu$ in Eq.~\ref{com3} is set to 0.3 if the joint CTC/attention simultaneous translation is used. The beam size is 10.

\subsection{Simultaneous ST results}
We compare our proposed simultaneous ST systems with the offline ST systems and conduct ablation studies to verify the effectiveness of our proposed CLE and CTC/attention joint translation methods. The experimental results on Fisher-CallHome Spanish are shown in Table~\ref{tab:fish}, where ASR encoder init. means to pre-train the encoder with ASR task \cite{inaguma-etal-2020-espnet}. The results show that the ASR pre-training works for both the offline and streaming models. In addition, 
the BLEU of our simultaneous ST system increases slightly with larger block size, while the latency (e.g. AL) also increases.
Furthermore, with our proposed Cross-lingual encoding method, the streaming Transformer ST model can choose joint CTC/attention simultaneous translation thus greatly outperforming vanilla attention-based decoding and achieving BLEU results close to that of offline systems.
\begin{table}[t]
  \caption{BLEU and average lagging (AL) (ms) of different ST systems on MUST-C En-De corpus.}
  \label{tab:must}
  \centering
  \setlength{\tabcolsep}{0.85mm}
  \begin{tabular}{l c c c | c c|c}
    \Xhline{3\arrayrulewidth}
    \multirow{2}{*}{Model}&
    \multirow{2}{*}{CLE} &{Block}&
    \multirow{2}{*}{Decode}&\multicolumn{2}{c|}{Must-C En-De}&\multirow{2}{*}{AL} \\
    \cline{5-6}
     & & Size & & COMMON&HE&  \\
    \hline
    Offline &\XSolidBrush&\XSolidBrush& Att&
    20.3&18.4&\multirow{2}{*}{---}\\
    { }{ } + ASR init.&\XSolidBrush&\XSolidBrush&Att&21.0&19.4&\\
    \hline
    Wait-5 \cite{ma-etal-2020-simulmt}  &&&&&\\
    $\cdot$ 440 ms step&\XSolidBrush&\XSolidBrush& Att&15.5&12.9&2896\\
    $\cdot$ 560 ms step&\XSolidBrush&\XSolidBrush& Att&16.1&14.6&3487\\
    \hline
    Streaming &\XSolidBrush &20& Att&14.2
    &12.1&3189\\
     { }{ } + ASR init.&\XSolidBrush &20& Att&15.8
     &13.4&2916\\
     Streaming &\Checkmark&20&Att&15.7
     &13.7&2872\\
    { }{ } + ASR init.&\Checkmark&20&Att&16.8
    &14.4&2844\\
    Streaming&\Checkmark&20&CTC/Att&19.3
    &17.0&\textbf{2484}\\
    { }{ } + ASR init.&\Checkmark&20&CTC/Att&\textbf{20.6}
    &\textbf{18.4}&2522\\
    \hline
    Streaming&\XSolidBrush &40& Att&15.4
    &13.2&3375\\
     { }{ } + ASR init.&\XSolidBrush &40& Att&17.4
     &14.7&3586\\
     Streaming&\Checkmark&40&Att&16.9
     &14.9&3351\\
    { }{ } + ASR init.&\Checkmark&40&Att&17.6
    &14.7&3361\\
    Streaming&\Checkmark&40&CTC/Att&20.4
    &19.4&\textbf{2998}\\
    { }{ } + ASR init.&\Checkmark&40&CTC/Att&\textbf{21.6}
    &\textbf{19.4}&3013\\
    \Xhline{3\arrayrulewidth}
  \end{tabular}
\end{table}

The experimental results on Must-C En-De are shown in Table~\ref{tab:must}, where offline and streaming respectively denote offline Transformer and streaming Transformer models, while Att and CTC/Att represent attention-based beam search and joint CTC/attention simultaneous translation, respectively.
The conclusions we get from Must-C En-De are similar to that of the Fisher-CallHome Spanish tasks: 1. The ASR initialization works for both offline and simultaneous ST systems; 2. The BLEU and latency of the simultaneous ST system increase with a larger block size being used; 3. With our proposed cross-lingual encoding method, the simultaneous ST model can further achieve significant improvement with our designed joint CTC/attention simultaneous translation. Furthermore, with our joint CTC/attention simultaneous translation, our simultaneous ST model greatly outperforms the wait-$k$ ($k$=5 here) prefix-to-prefix model \cite{ma-etal-2020-simulmt}. It should be noted that the wait-$k$ model also uses ASR pre-training initialization method.

\subsection{Streaming SLU results}
\begin{table}[t]
  \caption{The intent classification accuracy (\%) and endpoint latency (EP) (s) of different SLU systems on FSC corpus.}
  \label{tab:fsc}
  \centering
  \setlength{\tabcolsep}{1.2mm}
  \begin{tabular}{l c| c c| c c }
    \Xhline{3\arrayrulewidth}
    \multirow{2}{*}{Model}&
    {Block} &\multicolumn{2}{c}{Test}&\multicolumn{2}{c}{Dev} \\
    \cline{3-6}
     & {Size} &IC&EP&IC&EP \\
    \hline
    Offline Transformer & \XSolidBrush & 99.2 &--& 96.2&--\\
    \hline
    Streaming Transformer & 20 &55.3 &0.300 &55.5&\textbf{0.286}\\
    { }+ ASR intermediate loss & 20 & 56.8&\textbf{0.299} & 57.9&0.294\\
    \hline
    Streaming Transformer & 40 & 89.6& 0.348& 88.2&0.374\\
    { }+ ASR intermediate loss & 40 & 92.0&0.357 & 90.1&0.369\\
    \hline
    Streaming Transformer & 80 & 95.0 &0.379& 88.5&0.381\\
    { }+ ASR intermediate loss & 80  & \textbf{96.3}&0.392 & \textbf{90.7}&0.404\\
    \Xhline{3\arrayrulewidth}
  \end{tabular}
\end{table}
We compare our proposed streaming SLU systems with the offline SLU systems and conduct ablation studies to verify the effectiveness of the ASR-based intermediate loss regularization. The results are shown in Table~\ref{tab:fsc}, it should be noted that we predict the intent along with the ASR transcripts \cite{arora2021espnet}. The results show that the streaming SLU system can achieve classification performance close to that of the offline systems, which proves that predicting the intent based on previous chunks and then correcting it using beam search with future chunks works for streaming SLU tasks. In addition, after using the ASR-based intermediate loss, further improvement is achieved, which proves that it is beneficial to let the encoder first learn to distinguish tokens before trying to understand intents.

\begin{table}[t]
  \caption{The intent classification accuracy (\%) and endpoint latency (EP) (s) of different SLU systems on SLURP corpus.}
  \label{tab:slurp}
  \centering
  \setlength{\tabcolsep}{1.3mm}
  \begin{tabular}{l c| c c| c c}
    \Xhline{3\arrayrulewidth}
    \multirow{2}{*}{Model}&
    {Block} &\multicolumn{2}{c}{Test}&\multicolumn{2}{c}{Dev} \\
    \cline{3-6}
     & {Size} &IC&EP&IC&EP \\
    \hline
    Offline Transformer & \XSolidBrush & 84.7&-- & 85.2&--\\
    Streaming Transformer & 20 & 39.4 & \textbf{0.423}&39.5&\textbf{0.419}\\
    { }{ } + ASR CTC loss & 20 & 41.6&0.462 & 41.4&0.428\\
    \hline
    Streaming Transformer & 40 & 60.3&0.515&59.5&0.521\\
    { }{ } + ASR CTC loss & 40 &  61.1&0.504 & 60.2&0.508 \\
    \hline
    Streaming Transformer & 80 & 77.6&0.584&77.7&0.596\\
    { }{ } + ASR CTC loss & 80 &  \textbf{78.3}& 0.584& \textbf{78.6}& 0.595\\
    \Xhline{3\arrayrulewidth}
  \end{tabular}
\end{table}
We also conduct experiments on the SLURP corpus, and the results are shown in Table~\ref{tab:slurp}. We can see that the gap between the offline system and the streaming system decreases with the block size increases, although this comes at the cost of increased endpoint latency. Finally, with the help of ASR-based intermediate loss, we achieve 78.3\% classification accuracy which is close to that of the offline systems within an acceptable latency.

\section{Conclusions}

In this paper, we take the first step on applying the blockwise streaming Transformer developed for the ASR task to streaming SLU and simultaneous ST tasks.
To further improve the classification performance of the streaming SLU systems, we design an ASR-based intermediate loss, which encourages the encoder to first distinguish tokens before trying to understand intents.
As for the simultaneous ST task, we propose a cross-lingual encoding method by injecting a
CTC objective between encoder outputs and target translations.
In addition, the CTC is also employed to refine the search space and eliminate irregular alignments with the CTC prefix score, achieving joint CTC/attention simultaneous translation. 
Experimental results show that the blockwise streaming Transformer yields promising results on SLU and ST tasks,
especially with our ASR-based intermediate loss or joint CTC/attention simultaneous translation.
\bibliographystyle{IEEEtran}

\bibliography{mybib}


\end{document}